\documentclass[11pt,a4paper]{article}
\usepackage[margin=1in]{geometry}
\usepackage{amsmath,amssymb}
\usepackage{booktabs}
\usepackage{array}
\usepackage{multirow}
\usepackage{graphicx}
\graphicspath{{figures/}}
\usepackage{xcolor}
\usepackage{enumitem}
\usepackage{xspace}
\usepackage[colorlinks=true,allcolors=blue]{hyperref}

\newcommand{\ie}{i.e.\xspace}

\providecommand{\keywords}[1]{\par\medskip\noindent\textbf{Keywords:} #1}

\hypersetup{pdftitle={Annotation-Free Furniture Codes},
  pdfauthor={Benjamin Friedman}}

\title{Annotation-Free Furniture Codes:\\
What They Encode, and How Far They Transfer}
\author{Benjamin Friedman\\ DLR Group}
\date{}

\begin{document}
\maketitle

\begin{abstract}
Layout-based 3D scene synthesizers place each object using two human-annotated
channels: a categorical class label and a canonical-pose convention. We ask
whether a single self-supervised token derived from object geometry can replace
both, and study such tokens directly as a representation, decoupled from any
synthesizer. A Finite Scalar Quantization (FSQ) point-cloud autoencoder is
chamfer-trained on placed 3D-FUTURE furniture with no labels or pose annotations.
Diagnostic probes recover fine-category ($62.6\pm0.5$\,\%), super-category
($85.6\pm1.3$\,\%), and yaw ($52.7\pm0.5^\circ$) from the codes alone. Swapping
the chamfer target from the rotated to the un-rotated point cloud collapses the
yaw signal while raising class recovery, showing the codes' rotation content can
be set by the training objective. Scaling
across asset libraries needs codes that transfer; on an unseen dataset (ShapeNet),
alignment is category-dependent: box-like furniture transfers, organically-shaped
furniture does not, and a target-blind augmentation partly closes the gap.

\end{abstract}

\section{Introduction}
\label{sec:intro}

Layout-based indoor-scene synthesizers, autoregressive
(ATISS~\cite{paschalidou2021atiss}) and diffusion-based
(DiffuScene~\cite{tang2024diffuscene},
InstructScene~\cite{lin2024instructscene}) alike, predict per object a
categorical class index and an explicit yaw. Both are
supervision channels requiring human work: a class taxonomy someone must define
and assign, and a canonical-pose convention under which a predicted angle
$\theta$ orients different meshes consistently. That convention is subtle,
meaningful only if every mesh's local frame aligns to a shared semantic
``front,'' and neither the 3D-FUTURE~\cite{fu2021future} nor
3D-FRONT~\cite{fu20213dfront} papers document how it was set. Object geometry, by
contrast, is self-evident: point clouds are sampled from meshes with no human in
the loop.

A single self-supervised geometric token per object, a \emph{scene tokenizer}
standing in for the (class, angle) pair, is an attractive front-end, and one
that could scale across asset libraries without the per-library taxonomy work
labels demand. But a tokenizer is only as useful as what its tokens encode, a
question logically prior to any generator: what a token carries, what determines
it, and how far it survives a change of asset library. This paper characterizes
the tokens themselves, independent of any downstream generator; the
synthesizer is motivating context, not a deliverable here. We train a Finite
Scalar Quantization (FSQ)~\cite{mentzer2024fsq} point-cloud autoencoder with a
chamfer objective on placed (rotated) furniture point clouds (no class labels,
no pose annotations) and study the resulting 500-entry discrete codes.

\paragraph{Contributions.}
\begin{itemize}[leftmargin=1.5em,itemsep=2pt,topsep=2pt]
  \item \textbf{Probe quantification.} Small probes~\cite{alain2017linear}
  recover both super-/fine-category and yaw from the one-hot code, well above
  random and modal baselines, averaged over $n{=}3$ seed-and-split
  repetitions (Section~\ref{sec:encode}).
  \item \textbf{Loss-target control (headline).} A one-boolean intervention,
  replacing the chamfer target with the canonical (un-rotated) point cloud,
  collapses the yaw probe to modal baseline and raises class recovery. This
  quantifies, through the discrete bottleneck, that the rotation content is set
  by the \emph{supervision target}, and
  shows both rotation-aware and rotation-invariant regimes are reachable from one
  fixed pipeline (Section~\ref{sec:losstarget}). Only the rotated-target recipe
  is annotation-free; the canonical control uses the un-rotated mesh.
  \item \textbf{Cross-dataset code alignment (headline).} Encoding ShapeNet
  furniture through the frozen autoencoder, cross-dataset alignment is
  category-dependent: box-like categories transfer, organically-shaped ones do
  not (Section~\ref{sec:crossdataset}).
  \item \textbf{Domain-robust augmentation.} A generic digitization
  augmentation (with the target dataset never seen) partially closes the
  gap at zero within-dataset reconstruction cost (Section~\ref{sec:aug}).
\end{itemize}

We are explicit about scope. Every result here is a property of the codes.
We do \emph{not} train an end-to-end label-free synthesizer, report FID, or
study scene placement; those are separate downstream questions. The
contribution is a characterization of what a geometry-only shape vocabulary
contains and how it behaves under a loss-target change and a dataset shift.

\section{Related Work}
\label{sec:related}

\paragraph{Discrete 3D shape representations.}
VQ-VAEs~\cite{vandenoord2017vqvae} have been adapted to 3D in
AutoSDF~\cite{mittal2022autosdf}, ShapeFormer~\cite{yan2022shapeformer}, and
3DILG~\cite{zhang20223dilg}; MeshGPT~\cite{siddiqui2024meshgpt} learns a
triangle vocabulary. FSQ~\cite{mentzer2024fsq} removes the learned codebook
(no embedding table, no commitment loss, no EMA) via a fixed per-dimension
grid, which is why we adopt it; we include a VQ-VAE comparison
(supplementary). Point-BERT~\cite{yu2022pointbert} and
Point-MAE~\cite{pang2022pointmae} use masked, label-free pretraining on point
clouds; we share the geometry-only premise but target a
one-token-per-object vocabulary and analyse its contents directly.

\paragraph{Rotation in 3D representations.}
Two families dominate: equivariant architectures (Tensor Field
Networks~\cite{thomas2018tfn}, SE(3)-Transformers~\cite{fuchs2020se3}, Vector
Neurons~\cite{deng2021vn}) that bake in group structure, and learned
canonicalization (Canonical Capsules~\cite{sun2021canonical},
ConDor~\cite{sajnani2022condor}); Frame Averaging~\cite{puny2022fa} obtains
invariance by averaging any backbone over a frame. We ask a different question:
for a fixed non-equivariant pipeline, to what extent is the codes'
rotation-awareness set by the training objective? We isolate it to the loss
target (Section~\ref{sec:losstarget}).

\paragraph{Cross-dataset transfer.}
Whether a learned 3D representation transfers across asset libraries with
different tessellation, sampling, and modelling conventions is a practical
deployment question. We measure it directly in code space between
3D-FUTURE~\cite{fu2021future} and ShapeNet~\cite{chang2015shapenet}, and test
a standard robustness augmentation (jitter/dropout/voxel-snap) as a
domain-generalization~\cite{huang2021metasets} lever that never observes the target
dataset.

\section{Method}
\label{sec:method}

\begin{figure}[!t]
\centering
\includegraphics[width=\linewidth]{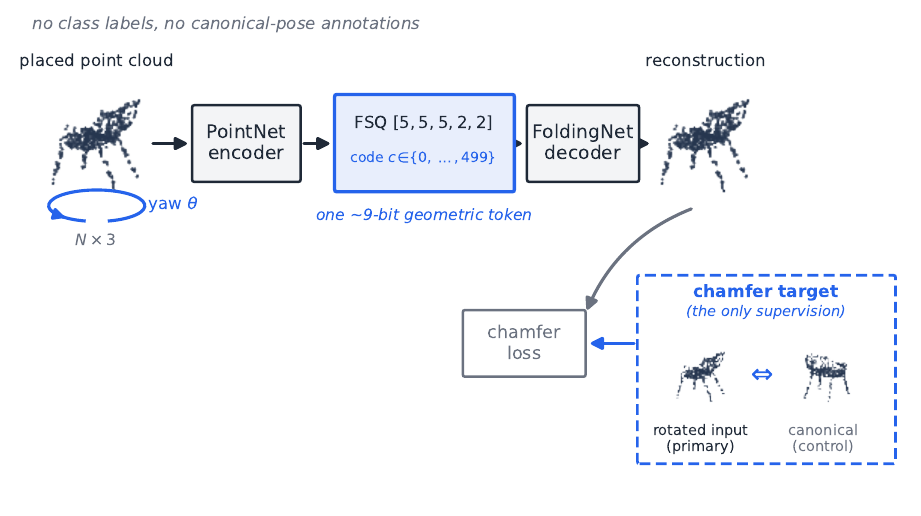}
\caption{Pipeline. A point cloud is encoded (PointNet), quantized by FSQ to a
single $500$-entry geometric code, and decoded (FoldingNet); the only training
signal is chamfer distance to a target. The \emph{chamfer target} is the one
knob we vary: the rotated input (primary recipe) or the canonical, un-rotated
point cloud (control, Section~\ref{sec:losstarget}). No class labels and no
canonical-pose annotations are consumed. (The token is ``${\sim}9$-bit'' in the
nominal sense: $500$ codes $\approx8.97$ bits; ${\sim}460$ are exercised
across rotated placements, $\approx8.85$ bits effective, and ${\sim}400$/$80\%$
over the canonical models alone; see supplementary.)}
\label{fig:pipeline}
\end{figure}

\subsection{Autoencoder and codebook}
A PointNet encoder~\cite{qi2017pointnet} maps an $N_{\text{in}}{\times}3$ input
point cloud
to a $128$-dim global feature, linearly projected to a $d_{\text{fsq}}{=}5$ FSQ latent,
$\tanh$-squashed and rounded to per-dimension levels $[5,5,5,2,2]$
($\prod_i L_i = 500$ codes), projected back to $128$ dims, and decoded by
FoldingNet~\cite{yang2018foldingnet} to an $N_{\text{out}}{\times}3$ output
cloud ($N_{\text{in}}{=}N_{\text{out}}{=}512$ in all experiments) (Fig~
\ref{fig:pipeline}). FSQ uses the
straight-through estimator; there is no learned codebook, commitment loss, or
EMA. Total parameters: $193$k. The training signal is geometry only: chamfer
distance between input and reconstruction. No class labels and no canonical
poses are consumed during training.

\subsection{Primary recipe and the loss-target control}
\label{sec:recipe}
Each 3D-FRONT placement carries a yaw $\theta$ about the vertical axis. Let
$\mathbf{P}_{\text{can}}$ be a mesh's canonical (un-rotated) point cloud and
$\mathbf{P}_\theta = R_y(\theta)\mathbf{P}_{\text{can}}$ its placed version.
\begin{description}[leftmargin=1.4em,style=nextline,itemsep=2pt,topsep=2pt]
\item[Primary recipe (rotated-target).] Input $\mathbf{P}_\theta$, target
$\mathbf{P}_\theta$: to reconstruct the placed cloud the encoder must encode
$\theta$. Strictly annotation-free in the chamfer loss.
\item[Canonical-target control.] Input $R_y(\Delta\theta)\mathbf{P}_\theta$,
target $\mathbf{P}_{\text{can}}$: the target is rotation-free, so the encoder
is asked to produce features that decode to the canonical pose regardless of
input rotation. This is the single controlled change ($\Delta\theta\in
[-180^\circ,180^\circ]$).
\end{description}
Architecture, optimizer, and codebook are held constant across the two; only
the chamfer target differs.

\subsection{Probe protocol}
\label{sec:probes}
To measure what the codes carry we train small probes~\cite{alain2017linear},
2-layer MLPs (hidden 64, dropout 0.1, Adam $3\mathrm{e}{-3}$, 30 epochs),
on a $500$-dim one-hot of the discrete code.
\begin{description}[leftmargin=1.4em,style=nextline,itemsep=2pt,topsep=2pt]
\item[P1 --- class.] 3D-FUTURE super-category (6 present) and fine-category
(27 with $\geq 30$ placements). Metrics: top-1/top-5, macro-F1. Baselines:
random and majority.
\item[P2 --- yaw.] A $(\cos\theta,\sin\theta)$ regressor (mean/median angular
error) and an $8$-bin classifier. Baselines: uniform-random ($90^\circ$) and
modal-yaw (${\sim}80^\circ$, from the strong axis-aligned prior). A
\emph{class-conditional} variant trains one regressor per super-category to
isolate within-class rotation beyond the class-yaw prior.
\end{description}
Split: 80/20 stratified by super-category, seed 42, held constant across
recipes. We use the taxonomy \emph{only} at evaluation (probe targets, purity);
the model never sees it.

\subsection{Cross-dataset alignment protocol}
\label{sec:align-protocol}
We measure whether foreign-dataset geometry lands where native geometry does.
For each ShapeNet~\cite{chang2015shapenet} category we sample
surface point clouds, normalize identically to the 3D-FUTURE canonical clouds,
encode them at $K\in\{0,90,180,270\}^\circ$ through the \emph{frozen}
autoencoder, and compute the pre-quantization embedding distance from each
ShapeNet cloud to the matching 3D-FUTURE super-category manifold, normalized
by the native intra-category spacing. We report the ratio
$r = d_{\text{ShapeNet}\to\text{native}} / d_{\text{native}}$; $r\approx 1$
means foreign geometry is indistinguishable from native
geometry, larger $r$ means off-manifold. Both terms are encoded with the same
checkpoint. $300$ clouds per category. $r$ is measured on the pre-quantization
latent, where distance is defined; the \emph{discrete codes} we deploy show the
same alignment (Section~\ref{sec:crossdataset}): a representation property,
not a latent artifact.

\section{Setup}
\label{sec:setup}
Furniture from 3D-FUTURE~\cite{fu2021future} ($8{,}229$ seen models,
${\sim}96$k placements), $N{=}512$ points. Adam, lr $10^{-3}$, batch $128$,
$15$ epochs, cosine annealing; validation chamfer on a held-out
$9{,}617$-placement split. Each configuration trains in $7$--$15$\,min on one
A100. Cross-dataset clouds are drawn from ShapeNetCore furniture synsets. Code,
configs, and per-run results will be released.

\section{What the codes encode}
\label{sec:encode}

\paragraph{Class identity.} Even with no label seen at training, per-code
super-category purity sits well above the $16.7\%$ random baseline and rises
monotonically with codebook size (a codebook-size sweep is in the
supplementary material). We anchor the study at 500 codes ($83.7\%$ purity),
a vocabulary a downstream model could plausibly learn to predict over.

\paragraph{Class and yaw, from the codes.} The probes
(Table~\ref{tab:probes}, ``rotated-target'' column) recover fine-category at
$62.6\pm0.5$\,\% (${\sim}4.4\times$ majority over $27$ classes) and super-category at
$85.6\pm1.3$\,\% (${\sim}5\times$ random); yaw follows at $52.7\pm0.5^\circ$ mean
angular error (${\sim}27^\circ$ better than modal). The codes carry both channels
a synthesizer would otherwise read from annotation.

\paragraph{Quantization cost.} To calibrate these numbers we probe the continuous
$5$-dim pre-quantization latent on the same three models: it recovers
super-category at $92.1\pm0.5\,\%$ and yaw at $45.4\pm0.3^\circ$, an effective
ceiling (the code is a deterministic function of that latent). Against the
\emph{discrete} probe on the same models (Table~\ref{tab:probes}: $85.6\,\%$,
$52.7^\circ$), quantizing to $500$ codes costs $6.5\,$pp super-category and
$7.3^\circ$ yaw: the codes retain most of the recoverable signal, and the
discrete-vs-continuous gap, not the absolute number, is the price of a compact
token.

\begin{figure}[!b]
\centering
\includegraphics[width=0.82\linewidth]{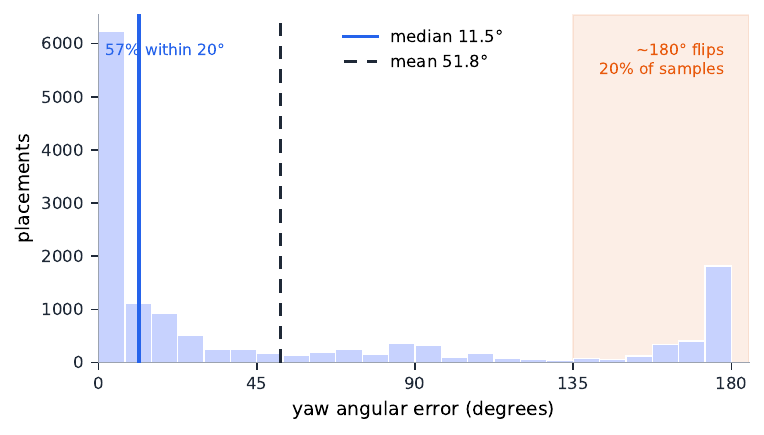}
\caption{Distribution of per-sample yaw angular error (rotated-target model, one
representative seed). Bimodal: a mode near $0^\circ$ ($57\,\%$ of placements
within $20^\circ$) and a second near $180^\circ$ ($20\,\%$, front/back flips
on symmetric furniture). The mean ($51.8^\circ$) is pulled up by the flip tail;
the median is $11.5^\circ$. (This seed; cf.\ the $n{=}3$ $52.7^\circ$ mean /
$10.2^\circ$ median in Table~\ref{tab:probes}, within noise.)}
\label{fig:yaw}
\end{figure}

\paragraph{The yaw error is bimodal.} The $52.7^\circ$ mean understates typical
accuracy: the median is $10.2^\circ$ and most placements are recovered within
$20^\circ$, but ${\sim}20\,\%$ suffer near-$180^\circ$ front/back flips on
symmetric furniture that drag the mean up (Fig~\ref{fig:yaw}). A geometry-only
code recovers yaw near-perfectly where orientation is unambiguous and fails where
the shape is symmetric: expected behaviour, not a defect.

\paragraph{The flips are front/back code collapse.} The bimodality has a
concrete mechanism. Encoding each mesh at $12$ yaws, we find the code is often
\emph{invariant} to a $180^\circ$ turn ($\mathrm{code}(\theta)=
\mathrm{code}(\theta{+}180^\circ)$) and this collapse is category-structured:
high for front/back-symmetric furniture (Pier/Stool $0.77$, Cabinet/Shelf/Desk
$0.67$, Table $0.64$) and low for clearly-oriented furniture (Sofa $0.14$, Bed
$0.13$, Chair $0.03$; symmetric objects also span far fewer distinct codes across
the $12$ yaws, ${\sim}5$ vs ${\sim}8$). A collapsed code cannot separate the two
orientations, so a code-based predictor guesses front/back at chance and errs by
${\sim}180^\circ$ on half of those placements; the mean collapse rate ($0.45$)
therefore predicts a ${\sim}0.22$ flip fraction, matching the ${\sim}20\%$ tail
in Fig~\ref{fig:yaw}. The flips are geometry, not noise: symmetric objects share
one code across their front and back placements.

\paragraph{Where the within-class rotation signal is strong.} The
class-conditional probe (Table~\ref{tab:cond}) shows the rotation signal is
strong for canonically-oriented classes (within-Chair $18.9^\circ$, Sofa
$18.6^\circ$, Bed $25.8^\circ$) and weak for Tables and Storage (within
${\sim}7^\circ$ of modal). This could reflect the classes' yaw
\emph{distributions} rather than their geometry: the oriented classes see more
varied training yaws in 3D-FRONT. A matched-yaw control (supplementary),
re-probing every class under an identical uniform yaw distribution, confirms the
split survives: it is geometric, not a distribution artifact (symmetric furniture
maps $\theta$ and $\theta{+}180^\circ$ to near-identical clouds, hence codes). The
control equalizes the \emph{evaluation} distribution, not the training one.

\begin{table}[!htbp]
\centering
\caption{Probe-based quantification, mean $\pm$ std over $n{=}3$
seed-and-split repetitions. Same architecture/optimizer/codebook; only the
chamfer target differs. Class: higher better; yaw error: lower better.}
\label{tab:probes}
\small
\begin{tabular}{l c c c}
\toprule
Probe / metric & Random / baseline & rotated-target & canonical-target \\
\midrule
\multicolumn{4}{l}{\textit{P1 --- class identity (code $\rightarrow$ class)}} \\
Super-cat top-1 (\%)       & 16.7 / maj.\ 36.5      & $85.6\pm1.3$      & $\mathbf{89.8\pm0.1}$ \\
Super-cat macro-F1         & 0.17                   & $0.745\pm0.027$   & $\mathbf{0.808\pm0.009}$ \\
Fine-cat top-1 (\%)        & 3.7 / maj.\ 14.2       & $62.6\pm0.5$      & $\mathbf{68.3\pm0.9}$ \\
Fine-cat top-5 (\%)        & 18.5                   & $96.3\pm0.3$      & $\mathbf{98.2\pm0.2}$ \\
\midrule
\multicolumn{4}{l}{\textit{P2 --- rotation (code $\rightarrow$ yaw, marginal)}} \\
Mean angular error (deg)   & 90 / modal 80          & $\mathbf{52.7\pm0.5}$ & $79.4\pm0.4$ \\
Median angular error (deg) & ---                    & $\mathbf{10.2\pm3.1}$ & $81.9\pm0.9$ \\
8-bin yaw top-1 (\%)       & 12.5 / modal 28.8      & $\mathbf{61.5\pm0.7}$ & $31.6\pm0.4$ \\
\bottomrule
\end{tabular}
\end{table}

\begin{table}[!htbp]
\centering
\caption{Class-conditional yaw probe: mean angular error (deg), lower better;
``$\Delta$ modal'' = improvement over the class-specific modal-yaw baseline
(positive = within-class rotation beyond the class label). Mean over $n{=}3$.}
\label{tab:cond}
\small
\begin{tabular}{l r r r r}
\toprule
\multirow{2}{*}{Super-cat} & \multicolumn{2}{c}{rotated-target} & \multicolumn{2}{c}{canonical-target} \\
\cmidrule(lr){2-3}\cmidrule(lr){4-5}
& probe & $\Delta$ modal & probe & $\Delta$ modal \\
\midrule
Chair   & $18.9\pm3.1$ & $\mathbf{+71.2}$ & $85.0\pm0.8$ & $+5.1$ \\
Bed     & $25.8\pm2.3$ & $\mathbf{+62.6}$ & $78.0\pm1.7$ & $+10.4$ \\
Sofa    & $18.6\pm1.1$ & $\mathbf{+72.6}$ & $84.4\pm1.3$ & $+6.8$ \\
Table   & $58.4\pm1.3$ & $+6.7$           & $66.9\pm1.3$ & $-1.9$ \\
Storage & $78.1\pm1.9$ & $+5.4$           & $83.1\pm1.1$ & $+0.5$ \\
Other   & $54.2\pm5.1$ & $-1.2$           & $55.6\pm2.6$ & $-2.6$ \\
\bottomrule
\end{tabular}
\end{table}

\section{The loss target controls the rotation encoding}
\label{sec:losstarget}

The primary recipe's codes encode rotation (Table~\ref{tab:probes}); where from?
Information-theoretically it is nearly forced: the encoder receives the
\emph{rotated} cloud in both regimes, so under a canonical (de-rotated) target
any retained orientation is penalised by chamfer, whereas a rotated target
rewards it. So the supervision target should govern the rotation content. We
quantify this through the discrete bottleneck
and show both regimes (rotation-aware and rotation-invariant) are reachable
from one fixed pipeline by flipping a single boolean, via the canonical-target
control (Section~\ref{sec:recipe}): architecture, optimizer, codebook, and split
are identical; only the chamfer target changes.

Two caveats. First, only the rotated-target recipe is annotation-free: the
canonical target requires the un-rotated mesh, which presupposes a human-defined
canonical-pose convention. The invariant regime is therefore a control, not a
second annotation-free recipe. Second, the annotation-free (rotated-target) codes
encode yaw \emph{entangled} with shape (revisited in the discussion).

\begin{figure}[!t]
\centering
\includegraphics[width=0.92\linewidth]{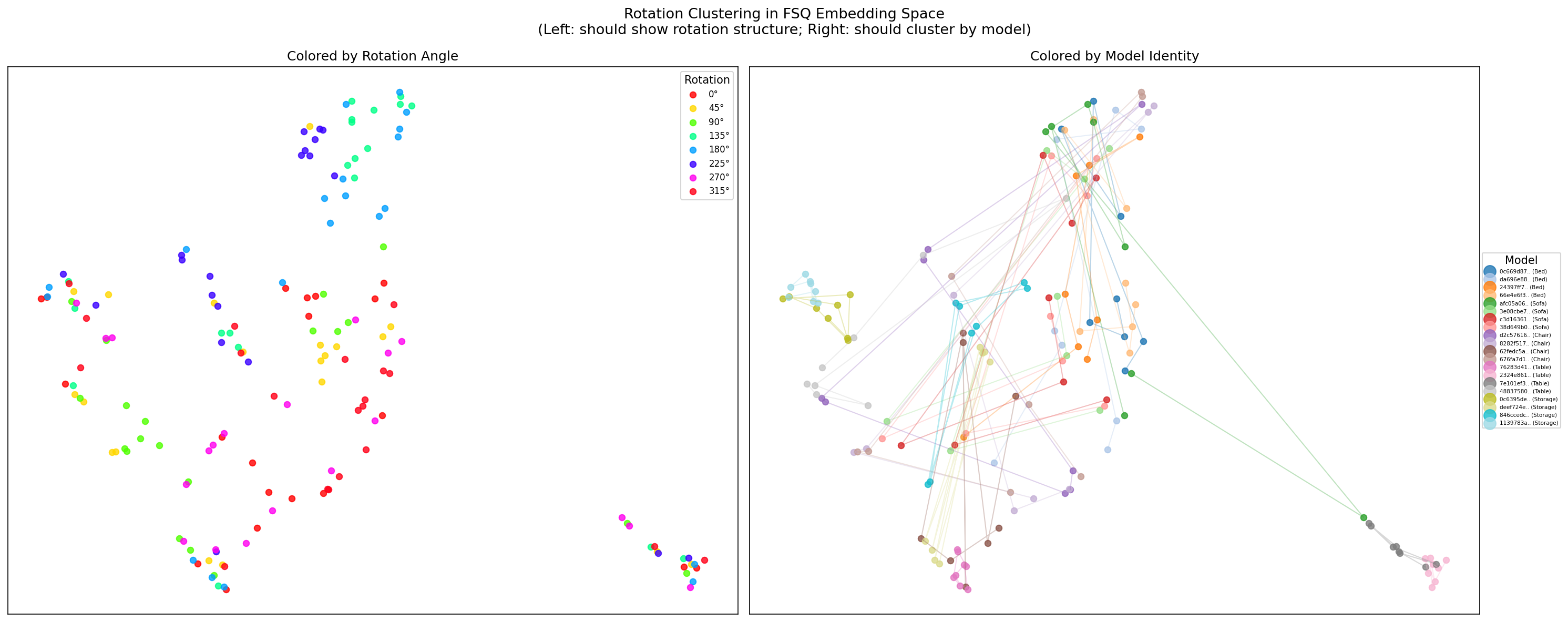}\\[2pt]
\includegraphics[width=0.92\linewidth]{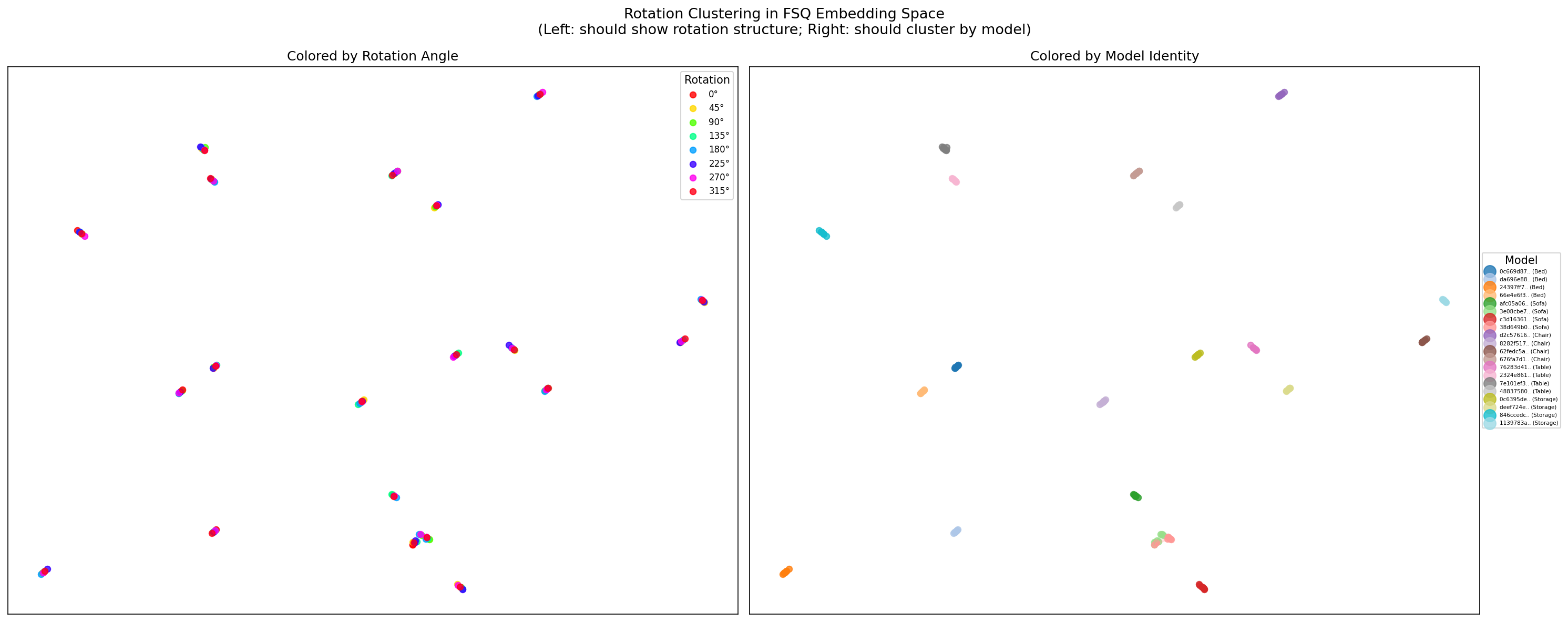}
\caption{Pre-quantization embeddings of held-out meshes, each encoded at $8$
yaws. \textbf{Top (rotated-target, primary):} points trace per-mesh rotation
\emph{arcs} (left, coloured by yaw) that also cluster by identity (right):
codes encode shape \emph{and} rotation. \textbf{Bottom (canonical-target,
control):} the arcs collapse to identity-only clusters: the same encoder,
same data, same held-out meshes; only the chamfer target differs. The rotation
encoding is set by the loss target.}
\label{fig:umap}
\end{figure}

\paragraph{The yaw encoding collapses.} Under the canonical target the yaw probe
drops to modal baseline (mean error $52.7^\circ\!\to\!79.4^\circ$, $8$-bin top-1
$61.5\%\!\to\!31.6\%$; Table~\ref{tab:probes}), and the class-conditional signal
zeroes out (Table~\ref{tab:cond}). The rotation encoding is therefore
\emph{data-induced}, a consequence of training against a rotation-bearing
target.

\paragraph{Freed capacity improves identity.} The same control \emph{raises}
class recovery: super-category top-1 $85.6\%\!\to\!89.8\%$ ($+4.2$\,pp), macro-F1
$0.745\!\to\!0.808$, fine-cat $62.6\%\!\to\!68.3\%$: capacity the primary
recipe spent on orientation is freed for identity, confirming it was being spent
on yaw. Reconstruction is unchanged within seed noise (chamfer
$0.0079\!\to\!0.0075$), so the collapse is not a fitting failure. The
training-distribution controls (supplementary) show this is robust to the input
augmentation: joint-augmentation controls (input and target rotated together) hold
reconstruction and utilization, while both canonical-target variants free
utilization.

\paragraph{Interpretation.} For the yaw-only 3D-FRONT setting, changing one
boolean in the loss switches the vocabulary between rotation-aware and
rotation-invariant: the behavioural change one would otherwise seek by
rebuilding the encoder to be equivariant, plus improved utilization. The
canonical-target regime is best read as a learned
canonicalization~\cite{sun2021canonical,sajnani2022condor} with a discrete
bottleneck; we propose no new mechanism (the canonical pose is given by the
dataset). The invariance is empirical, holding over the training augmentation's
support (yaw only, not full SO(3)), where an equivariant
encoder~\cite{thomas2018tfn,fuchs2020se3,deng2021vn} or frame
averaging~\cite{puny2022fa} would generalize by construction.

\section{Cross-dataset code alignment}
\label{sec:crossdataset}

Does a geometry-only vocabulary transfer to a \emph{different} asset library?
We encode ShapeNet furniture through the frozen 500-code autoencoder and
measure per-category alignment to the native 3D-FUTURE manifold
(Section~\ref{sec:align-protocol}). The answer is category-dependent
(Table~\ref{tab:cross}): box-like categories (cabinet, bookshelf, table,
lamp) already align ($r=0.9$--$1.6\times$ native spacing), whereas
organically-shaped categories (bed, sofa, chair) sit $3.2$--$5.1\times$ off
the manifold. Five confounds were ruled out:
discrete-vs-continuous latent, mesh quality, pool size (a size sweep is
invariant), orientation (both datasets Y-up; extents matched), and category
(measured across all seven).

\begin{figure}[t]
\centering
\includegraphics[width=0.72\linewidth]{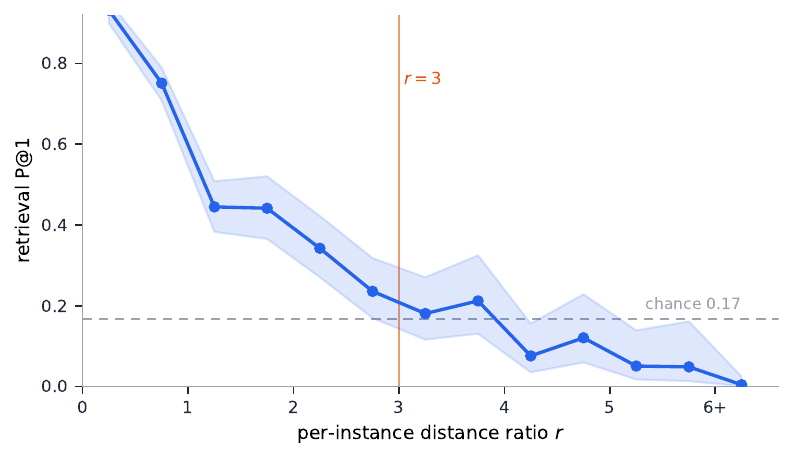}
\caption{The cross-dataset transfer curve. Each of $2{,}033$ ShapeNet query
meshes is scored by its own distance ratio $r$ to the native manifold and binned;
$y$ is the fraction whose nearest native neighbour in code space is the correct
super-category (P@1; band is Wilson $95\%$). Retrieval degrades smoothly with
$r$ (a gradient, not a step) and reaches chance ($0.17$) near $r\approx3$.}
\label{fig:curve}
\end{figure}

\paragraph{A continuous transfer gradient.} $r$ is a manifold-distance
ratio; we tie it to a task by retrieving each ShapeNet query's nearest native
neighbours in code space (balanced 6-way gallery, chance $0.17$) and asking
whether the nearest is the correct super-category (Table~\ref{tab:cross}, P@1).
At the category level every $r<3$ category retrieves correctly $3$--$5\times$
above chance ($0.55$--$0.78$), while sofa and chair fall to or below it ($0.07$,
$0.09$); the two rankings agree (Spearman $-0.79$). Scoring each of the
$2{,}033$ query meshes by its \emph{own} distance ratio resolves the shape
(Fig~\ref{fig:curve}): P@1 declines smoothly and monotonically from ${\sim}0.8$
at $r{<}1$ to near zero at $r{>}6$, with no cliff, passing through chance around
$r\approx3$. The $r<3$ cut is therefore best read not as a hard boundary but as
the point where geometry-only retrieval decays to chance; transfer is a
continuum, and the binary label a convenience: an $r{=}1$ query retrieves far
better than an $r{=}2.5$ one, though both nominally ``transfer.'' This anchor is
at the \emph{super-category} level: $r$ tracks whether a foreign query lands among
the right kind of native furniture, not that it retrieves the right individual
shape (fine-grained retrieval is left open).

\begin{table}[!htbp]
\centering
\caption{Cross-dataset code alignment. $r = $ ShapeNet-to-native distance
$/$ native intra-category spacing, encode-native matched methodology, 300
clouds/category, 500-code baseline autoencoder. $r\approx 1$: foreign geometry
indistinguishable from native. \textbf{P@1}: fraction of ShapeNet
queries whose nearest native neighbour in the pre-quantization embedding is the
correct super-category, over a balanced 6-way gallery (chance $0.17$); it anchors $r$ to
a retrieval task. Spearman($r$, P@1)$=-0.79$. The transfers/gap column is a
two-way discretization of a continuum: retrieval degrades \emph{smoothly} with
$r$ (Fig~\ref{fig:curve}), so the labels are a reading aid, not a hard
boundary.}
\label{tab:cross}
\small
\begin{tabular}{l l r r l}
\toprule
ShapeNet cat & Native target & $r$ & P@1 & verdict \\
\midrule
cabinet   & Cabinet/Shelf/Desk & 0.9$\times$ & 0.60 & transfers \\
bookshelf & Cabinet/Shelf/Desk & 1.0$\times$ & 0.78 & transfers \\
table     & Table              & 1.5$\times$ & 0.55 & transfers \\
lamp      & Others             & 1.6$\times$ & 0.72 & transfers \\
\midrule
bed       & Bed                & 3.2$\times$ & 0.36 & gap \\
sofa      & Sofa               & 3.9$\times$ & 0.07 & gap \\
chair     & Chair              & 5.1$\times$ & 0.09 & gap \\
\bottomrule
\end{tabular}
\end{table}

\paragraph{The discrete codes agree with the embedding.} $r$ and P@1 are defined
on the continuous pre-quantization embedding, but the paper's unit of study is
the code, so we repeat the alignment on the discrete index. Predicting each
ShapeNet query's native super-category from its code alone (majority native vote
over the balanced gallery) gives a discrete P@1 that tracks the embedding P@1
across categories (Spearman $0.93$) and $r$ (Spearman $-0.75$): transferring
categories score $0.46$--$0.61$, sofa and chair fall to chance ($0.08$, $0.05$).
Independently, each transferring category's $500$-bin code histogram most
resembles the correct native super-category, whereas sofa and chair's do not. Because the discrete codes reproduce the alignment measured on the
continuous latent, the cross-dataset result is a property of the codes
themselves, not an artifact of measuring the pre-quantization embedding.

\paragraph{Interpretation.} The split tracks geometric stereotypy: box-like
furniture is nearly identical in silhouette across libraries (a cabinet is a
cuboid everywhere), so its codes are dataset-agnostic; chairs, sofas, and beds
vary far more, and that variation is what a geometry-only code is sensitive to. A
label-based vocabulary is more robust to this shift, but not for free: carrying
labels across datasets means reconciling two taxonomies of differing coverage and
specificity (does ``chair'' map to one bucket, or split across armchair / stool /
dining-chair?), itself manual work. The contrast is thus a trade: geometry
codes need no taxonomy alignment but drift off-manifold where shape varies most,
while labels resist that drift only once a human has aligned the label spaces. We
do not evaluate downstream retrieval or placement here.

\section{A generic augmentation partially closes the gap}
\label{sec:aug}

Can the gap shrink without letting the encoder see the target dataset, \ie
preserving a genuine unseen-dataset test? We retrain on 3D-FUTURE with a generic
\emph{digitization} augmentation (denoising: input augmented, target clean) ---
coordinate jitter, random point dropout, and voxel-snap (a low-poly/tessellation
proxy). The menu is standard 3D robustness augmentation, justified generically
rather than tuned to ShapeNet; the encoder never sees ShapeNet.

\begin{figure}[!t]
\centering
\includegraphics[width=\linewidth]{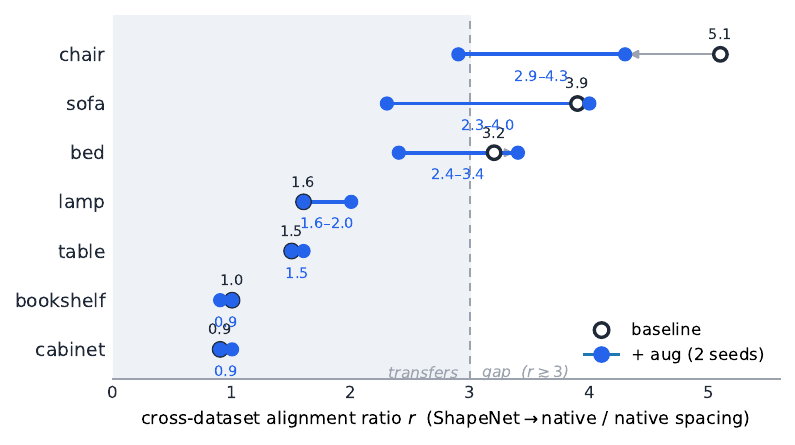}
\caption{Cross-dataset alignment ratio $r$ per ShapeNet category on the
$500$-code rotated-target model (open = baseline; filled = $+$digitization
augmentation, the two training seeds joined by a bar; gray arrow = baseline
$\to$aug shift). $r<3$ (shaded) is on-manifold. Box-like categories transfer
robustly and are seed-stable; the shape-variable categories (bed, sofa, chair)
shrink toward the boundary at zero within-dataset cost, but their two aug seeds
\emph{straddle} it; the per-category effect is seed-dependent
(Table~\ref{tab:aug}).}
\label{fig:cross}
\end{figure}

The augmentation gives \emph{partial} domain robustness at zero within-dataset
cost (Table~\ref{tab:aug}): validation chamfer is unchanged ($0.0072$ vs the
$0.0073$ baseline), and the already-transferring box-like categories stay put
across both training seeds. For the shape-variable categories the effect is real
but \emph{seed-dependent}: augmentation pulls sofa, bed, and chair from their
$3.2$--$5.1\times$ baselines down toward the $r{=}3$ boundary, but two
augmentation seeds disagree on which of them crosses it (sofa $2.3$/$4.0$, bed
$3.4$/$2.4$, chair $4.3$/$2.9$). We therefore report the aggregate shift toward
transfer, not a per-category ordering.

\begin{table}[!htbp]
\centering
\caption{Domain-robust digitization augmentation vs the baseline, same
encode-native diagnostic. Augmented VAEs never observe ShapeNet;
within-3D-FUTURE val chamfer $0.0072 \approx 0.0073$ baseline (no regression).
All are $500$-code rotated-target models in the same regime, differing only in
the augmentation. \textbf{s1}, \textbf{s2} are two augmentation training seeds:
box-like categories transfer robustly across both; the shape-variable categories
(sofa/bed/chair) are seed-variable, hovering near the $r{=}3$ boundary.}
\label{tab:aug}
\small
\begin{tabular}{l r r r l}
\toprule
Category & baseline $r$ & aug $r$ (s1) & aug $r$ (s2) & note \\
\midrule
cabinet   & 0.9$\times$ & 0.9$\times$ & 1.0$\times$ & transfers (stable) \\
bookshelf & 1.0$\times$ & 0.9$\times$ & 1.0$\times$ & transfers (stable) \\
table     & 1.5$\times$ & 1.5$\times$ & 1.6$\times$ & transfers (stable) \\
lamp      & 1.6$\times$ & 2.0$\times$ & 1.6$\times$ & transfers (stable) \\
\midrule
sofa      & 3.9$\times$ & 2.3$\times$ & 4.0$\times$ & seed-variable \\
bed       & 3.2$\times$ & 3.4$\times$ & 2.4$\times$ & seed-variable \\
chair     & 5.1$\times$ & 4.3$\times$ & 2.9$\times$ & seed-variable \\
\bottomrule
\end{tabular}
\end{table}

\paragraph{Reading.} Style-level augmentation removes the part of the
cross-dataset gap attributable to digitization differences (sampling density,
tessellation, scan noise). Its effect on the box-like categories is null (they
already transfer) and stable across seeds; on the three shape-variable categories
it shrinks the gap on average but with enough seed variance that no single one
reliably crosses $r{<}3$. This is consistent with a residual, genuine silhouette
shift that surface-style augmentation only partly removes, and points to
multi-source training with a third held-out dataset as the honest next lever
(deferred). We deliberately exclude training the encoder on ShapeNet: even
with chairs held out, it would let the encoder adapt to ShapeNet style through
the other categories, silently degrading a dataset-level unseen test to a
class-level one.

\section{Discussion and limitations}
\label{sec:limitations}

\paragraph{What these results are.} A characterization of a geometry-only shape
vocabulary: it carries class and yaw recoverably; its rotation content is set by
the loss target; and it transfers across datasets in a
category-dependent way a generic augmentation partly repairs. The probe protocol
(${\sim}90$k pre-extracted codes, under two minutes on one GPU) is cheap enough
to serve as a tokenizer-quality unit test before any downstream use.

\paragraph{Bearing on scene generation.} We build no scene synthesizer, but both
findings bear on one. A generator built on these tokens reads each in place of
the (class, pose) annotations current layout-based models take per object, so the class and
orientation it can condition on are bounded by what the token encodes, which our
probes measure (Sections~\ref{sec:encode}--\ref{sec:losstarget}); the loss-target
control makes that bound a design choice, not a fixed encoder property. The cross-dataset result (Section~\ref{sec:crossdataset}) is about
\emph{scale}: a multi-library generator would reuse one geometry-only vocabulary
across asset libraries, and $r$ marks per category where that reuse holds, and
where it would silently degrade.

\paragraph{Limitations.}
\begin{itemize}[leftmargin=1.4em,itemsep=2pt,topsep=2pt]
\item \textbf{Downstream out of scope.} We show what the codes contain and how
they transfer, not that a synthesizer built on them matches a label-supervised
one; probes measure recoverability \emph{from the code}, not predictability from
autoregressive context.
\item \textbf{Entangled pose and shape.} The primary recipe packs yaw and shape
into one ${\sim}9$-bit token, so pose is not independently addressable:
re-orienting an object means jumping to a different code that may also change its
shape: a genuine \emph{design} problem for a layout-editable generator, not
just an unrun experiment. The arcs in Fig~\ref{fig:umap} show yaw varies smoothly
in the latent, but not that a controllable, globally factorized yaw axis exists.
The canonical-target regime removes the entanglement but is not annotation-free.
\item \textbf{Single dataset pair; one VAE.} Alignment is measured 3D-FUTURE
$\leftrightarrow$ ShapeNet with one 500-code autoencoder; the augmentation is a
single configuration (two seeds), seed-variable per-category on the
shape-variable classes (Table~\ref{tab:aug}).
\item \textbf{Yaw only.} 3D-FRONT placements rotate about the vertical axis; the
loss-target control speaks to yaw, not full SO(3).
\item \textbf{Within-class rotation confound (training-side).} The matched-yaw
control (supplementary) equalizes the \emph{evaluation} yaw distribution but not
the training one, so geometric observability and training-yaw variety are not
fully separated (that would need retraining on yaw-balanced data).
\item \textbf{No external calibration point.} We report recoverability in absolute
terms. Established SSL point encoders (e.g.\ Point-BERT /
Point-MAE~\cite{yu2022pointbert,pang2022pointmae}) are not drop-in baselines
(per-patch, multi-token, and far higher-capacity than our single ${\sim}9$-bit
code), so a fair use is a clearly-labeled capacity-mismatched ceiling (a pooled
SSL embedding, probed), left to future work.
\item \textbf{Uneven seed coverage.} The core probe table (Table~\ref{tab:probes})
carries $n{=}3$ error bars; the follow-ups --- continuous-latent reference,
yaw-error distribution (Fig~\ref{fig:yaw}), retrieval P@1, and transfer curve
(Fig~\ref{fig:curve}, Wilson intervals over $2{,}033$ meshes) --- are
single-seed, as are the VQ and encoder-backbone comparisons.
\end{itemize}

\section{Conclusion}
\label{sec:conclusion}

A chamfer-trained FSQ autoencoder, given no class labels and no canonical-pose
annotations, produces codes from which a small probe recovers class
(fine-category $62.6\pm0.5$\,\%, super-category $85.6\pm1.3$\,\%) and yaw
($52.7\pm0.5^\circ$): the two channels synthesizers read from annotation. Swapping the chamfer target (rotated $\rightarrow$
canonical) collapses the yaw encoding to modal baseline and raises class
recovery, so rotation content is set by the objective.
Across datasets alignment is category-dependent (box-like transfers;
bed/sofa/chair sit $3.2$--$5.1\times$ off-manifold), which a target-blind
augmentation partly and seed-dependently closes. Scope is narrow (yaw only,
one dataset pair, one autoencoder), and a label-free synthesizer on these codes
remains future work.



\clearpage
\appendix
\setcounter{section}{0}
\setcounter{table}{0}
\setcounter{figure}{0}
\renewcommand{\thesection}{S\arabic{section}}
\renewcommand{\thetable}{S\arabic{table}}
\renewcommand{\thefigure}{S\arabic{figure}}

\section{Supporting analyses}

\paragraph{Codebook-size sweep.} Weighted super-category purity rises
monotonically with codebook size ($76.1\%$ at 108 codes to $89.1\%$ at
$3{,}125$, all far above the $16.7\%$ random baseline), while chamfer plateaus
past 500 codes and the effective vocabulary scales sub-linearly
(Table~\ref{tab:v1}). We anchor the main study at 500 codes as a size a
downstream model could plausibly learn to predict over.

\begin{table}[!htbp]
\centering
\caption{Codebook-size sweep. Effective vocabulary (used codes) scales
sub-linearly; chamfer plateaus past 500 codes; weighted super-category purity
improves monotonically (random $\approx 16.7\%$). \emph{Used/Util count distinct
codes over the canonical (unrotated) models; rotation exercises more: the
$500$-code baseline uses $\sim$460 ($\sim$93\%) over rotated placements, the set
the probes encode.}}
\label{tab:v1}
\small
\begin{tabular}{lrrrrr}
\toprule
Config & Codes & Used & Util & Chamfer & Sup.\ Purity \\
\midrule
108-code $(3,3,3,2,2)$             & 108   & 106   & 98.1\,\% & 0.0097 & 76.1\,\% \\
500-code $(5,5,5,2,2)$ [baseline]  & 500   & 400   & 80.0\,\% & 0.0079 & 83.7\,\% \\
1250-code $(5,5,5,5,2)$            & 1{,}250 & 732 & 58.6\,\% & 0.0077 & 86.4\,\% \\
3125-code $(5,5,5,5,5)$            & 3{,}125 & 1{,}235 & 39.5\,\% & 0.0069 & 89.1\,\% \\
\bottomrule
\end{tabular}
\end{table}

\paragraph{Bottleneck: FSQ vs VQ-VAE.} At a matched 500-code budget, an
out-of-the-box VQ-VAE~\cite{vandenoord2017vqvae} (EMA codebook, commitment
$0.25$, single seed) collapses to $7.5$\,\% utilization (36 codes) under the
rotated target vs FSQ's $96.3$\,\% ($481\pm8$; both measured over rotated
placements in this matched-budget comparison, consistent with the baseline's
$93\%$ over placements and $80\%$ over canonical models, Table~\ref{tab:v1}); the
collapse propagates to every probe (super-cat top-1 $68.4$ vs $85.6$). Under the canonical target,
where the task needs less capacity, the gap narrows (utilization $22.2$ vs
$84.3$\,\%; super-cat $86.8$ vs $89.8$). The loss-target rotation finding
survives qualitatively under VQ (more rotation under rotated-target, more class
under canonical), just at lower magnitudes, confirming it is a property of the
objective, not the bottleneck. We do not claim the utilization gap as a result:
codebook collapse is a well-documented VQ failure mode with well-known
fixes (codebook reset~\cite{dhariwal2020jukebox}, k-means
init~\cite{lancucki2020robust}, lower commitment), and these are single-seed, out-of-the-box numbers. Read this only
as a practitioner note (FSQ gave us high utilization with no such tuning),
not as evidence that FSQ is fundamentally better.

\paragraph{Encoder backbone.} Swapping PointNet for DGCNN~\cite{wang2019dgcnn} at $k{=}20$ leaves
chamfer, utilization, and neighbourhood purity within run-to-run noise
(chamfer $0.0079$ vs $0.0076$; utilization $80$ vs $83$\,\%); $k{=}10$
underfits. Both encoders are non-equivariant and max-pool over $512$ points,
which discards much of the local edge structure EdgeConv exposes; the
comparison may differ at larger point counts. This is one seed: we report no
evidence of a difference at this scale, not equivalence, which a single seed
cannot establish.

\paragraph{Matched-yaw control.} The within-class rotation signal (main paper,
Section~\ref{sec:encode}) is stronger for canonically-oriented classes, but those
classes also see more varied training yaws in 3D-FRONT. To separate geometry from
distribution we re-probe every class under an \emph{identical} uniform yaw
distribution: each mesh rotated through $24$ evenly-spaced yaws, mesh-disjoint
eval. The split persists (Table~\ref{tab:matchedyaw}): Sofa/Chair/Bed recover
orientation at $31$--$43^\circ$ (uniform baseline $90^\circ$), while
Table/Others/Cabinet/Stool sit at $79$--$88^\circ$, essentially baseline. With the
distribution matched the difference is geometric: front/back-symmetric furniture
maps $\theta$ and $\theta{+}180^\circ$ to near-identical clouds (hence codes),
so orientation is unrecoverable regardless of training. The control equalizes the
\emph{evaluation} distribution; the training distribution stays natural (fully
removing that confound would need retraining on yaw-balanced data).

\begin{table}[!htbp]
\centering
\caption{Matched-yaw control. Class-conditional yaw error (mean/median deg,
lower better) with every class given an \emph{identical} uniform yaw
distribution ($24$ yaws/mesh, mesh-disjoint eval; uniform baseline $90^\circ$).
The strong/weak split of Table~\ref{tab:cond} (main paper) survives distribution
matching, isolating it to geometric observability rather than yaw variety.}
\label{tab:matchedyaw}
\small
\begin{tabular}{l r r}
\toprule
Super-cat & Matched mean & Matched median \\
\midrule
Sofa               & 31.5 & 18.1 \\
Chair              & 32.1 & 24.7 \\
Bed                & 43.3 & 25.3 \\
\midrule
Table              & 79.1 & 71.3 \\
Others             & 84.1 & 79.0 \\
Cabinet/Shelf/Desk & 88.1 & 86.5 \\
Pier/Stool         & 88.3 & 88.1 \\
\bottomrule
\end{tabular}
\end{table}

\paragraph{Training-distribution controls.} Table~\ref{tab:v6} gives the
training-time chamfer and utilization for the primary recipe and the four
augmentation/target controls discussed in Section~\ref{sec:losstarget} of the
main paper.

\begin{table}[!htbp]
\centering
\caption{Primary recipe (row 1) and four training-distribution controls at 500
codes. Rows 2--3 vary the input distribution (target $=$ rotated input); rows
4--5 vary the supervision target (canonical). Chamfer/utilization are
training-time; recoverability is in Table~\ref{tab:probes} (main paper).}
\label{tab:v6}
\small
\begin{tabular}{l l r r}
\toprule
Aug & Target & Chamfer & Util \\
\midrule
--- (baseline)        & rotated   & 0.0079 & 80.0\,\% \\
$\pm15^\circ$ joint   & rotated   & 0.0080 & 82.0\,\% \\
$\pm180^\circ$ joint  & rotated   & 0.0089 & 66.4\,\% \\
$\pm15^\circ$         & canonical & 0.0075 & 79.6\,\% \\
$\pm180^\circ$        & canonical & 0.0075 & 87.6\,\% \\
\bottomrule
\end{tabular}
\end{table}



\begin{thebibliography}{99}
\small

\bibitem{mentzer2024fsq}
F.~Mentzer, D.~Minnen, E.~Agustsson, and M.~Tschannen.
\newblock Finite scalar quantization: VQ-VAE made simple.
\newblock In \emph{ICLR}, 2024.

\bibitem{vandenoord2017vqvae}
A.~van den Oord, O.~Vinyals, and K.~Kavukcuoglu.
\newblock Neural discrete representation learning.
\newblock In \emph{NeurIPS}, 2017.

\bibitem{paschalidou2021atiss}
D.~Paschalidou, A.~Kar, M.~Shugrina, K.~Kreis, A.~Geiger, and S.~Fidler.
\newblock ATISS: Autoregressive transformers for indoor scene synthesis.
\newblock In \emph{NeurIPS}, 2021.

\bibitem{tang2024diffuscene}
J.~Tang, Y.~Nie, L.~Markhasin, A.~Dai, J.~Thies, and M.~Nie\ss{}ner.
\newblock DiffuScene: Denoising diffusion models for generative indoor scene
synthesis.
\newblock In \emph{CVPR}, 2024.

\bibitem{lin2024instructscene}
C.~Lin and Y.~Mu.
\newblock InstructScene: Instruction-driven 3D indoor scene synthesis with
semantic graph prior.
\newblock In \emph{ICLR}, 2024.

\bibitem{fu2021future}
H.~Fu, R.~Jia, L.~Gao, M.~Gong, B.~Zhao, S.~Maybank, and D.~Tao.
\newblock 3D-FUTURE: 3D furniture shape with texture.
\newblock \emph{IJCV}, 2021.

\bibitem{fu20213dfront}
H.~Fu, B.~Cai, L.~Gao, L.-X.~Zhang, J.~Wang, C.~Li, Q.~Zeng, C.~Sun, R.~Jia,
B.~Zhao, and H.~Zhang.
\newblock 3D-FRONT: 3D furnished rooms with layouts and semantics.
\newblock In \emph{ICCV}, 2021.

\bibitem{chang2015shapenet}
A.~X.~Chang, T.~Funkhouser, L.~Guibas, P.~Hanrahan, Q.~Huang, Z.~Li,
S.~Savarese, M.~Savva, S.~Song, H.~Su, J.~Xiao, L.~Yi, and F.~Yu.
\newblock ShapeNet: An information-rich 3D model repository.
\newblock \emph{arXiv:1512.03012}, 2015.

\bibitem{alain2017linear}
G.~Alain and Y.~Bengio.
\newblock Understanding intermediate layers using linear classifier probes.
\newblock In \emph{ICLR Workshop}, 2017.

\bibitem{qi2017pointnet}
C.~R.~Qi, H.~Su, K.~Mo, and L.~J.~Guibas.
\newblock PointNet: Deep learning on point sets for 3D classification and
segmentation.
\newblock In \emph{CVPR}, 2017.

\bibitem{yang2018foldingnet}
Y.~Yang, C.~Feng, Y.~Shen, and D.~Tian.
\newblock FoldingNet: Point cloud auto-encoder via deep grid deformation.
\newblock In \emph{CVPR}, 2018.

\bibitem{mittal2022autosdf}
P.~Mittal, Y.-C.~Cheng, M.~Singh, and S.~Tulsiani.
\newblock AutoSDF: Shape priors for 3D completion, reconstruction and
generation.
\newblock In \emph{CVPR}, 2022.

\bibitem{yan2022shapeformer}
X.~Yan, L.~Lin, N.~J.~Mitra, D.~Lischinski, D.~Cohen-Or, and H.~Huang.
\newblock ShapeFormer: Transformer-based shape completion via sparse
representation.
\newblock In \emph{CVPR}, 2022.

\bibitem{zhang20223dilg}
B.~Zhang, M.~Nie\ss{}ner, and P.~Wonka.
\newblock 3DILG: Irregular latent grids for 3D generative modeling.
\newblock In \emph{NeurIPS}, 2022.

\bibitem{siddiqui2024meshgpt}
Y.~Siddiqui, A.~Alliegro, A.~Artemov, T.~Tommasi, D.~Sirigatti, V.~Rosov,
A.~Dai, and M.~Nie\ss{}ner.
\newblock MeshGPT: Generating triangle meshes with decoder-only transformers.
\newblock In \emph{CVPR}, 2024.

\bibitem{yu2022pointbert}
X.~Yu, L.~Tang, Y.~Rao, T.~Huang, J.~Zhou, and J.~Lu.
\newblock Point-BERT: Pre-training 3D point cloud transformers with masked
point modeling.
\newblock In \emph{CVPR}, 2022.

\bibitem{pang2022pointmae}
Y.~Pang, W.~Wang, F.~E.~H.~Tay, W.~Liu, Y.~Tian, and L.~Yuan.
\newblock Masked autoencoders for point cloud self-supervised learning.
\newblock In \emph{ECCV}, 2022.

\bibitem{thomas2018tfn}
N.~Thomas, T.~Smidt, S.~Kearnes, L.~Yang, L.~Li, K.~Kohlhoff, and P.~Riley.
\newblock Tensor field networks: Rotation- and translation-equivariant neural
networks for 3D point clouds.
\newblock \emph{arXiv:1802.08219}, 2018.

\bibitem{fuchs2020se3}
F.~Fuchs, D.~Worrall, V.~Fischer, and M.~Welling.
\newblock SE(3)-Transformers: 3D roto-translation equivariant attention
networks.
\newblock In \emph{NeurIPS}, 2020.

\bibitem{deng2021vn}
C.~Deng, O.~Litany, Y.~Duan, A.~Poulenard, A.~Tagliasacchi, and L.~Guibas.
\newblock Vector neurons: A general framework for SO(3)-equivariant networks.
\newblock In \emph{ICCV}, 2021.

\bibitem{sun2021canonical}
W.~Sun, A.~Tagliasacchi, B.~Deng, S.~Sabour, S.~Yazdani, G.~Hinton, and
K.~M.~Yi.
\newblock Canonical capsules: Self-supervised capsules in canonical pose.
\newblock In \emph{NeurIPS}, 2021.

\bibitem{sajnani2022condor}
R.~Sajnani, A.~Poulenard, J.~Jain, R.~Dua, L.~Guibas, and S.~Sridhar.
\newblock ConDor: Self-supervised canonicalization of 3D pose for partial
shapes.
\newblock In \emph{CVPR}, 2022.

\bibitem{puny2022fa}
O.~Puny, M.~Atzmon, H.~Ben-Hamu, I.~Misra, A.~Grover, E.~J.~Smith, and
Y.~Lipman.
\newblock Frame averaging for invariant and equivariant network design.
\newblock In \emph{ICLR}, 2022.

\bibitem{huang2021metasets}
C.~Huang, Z.~Cao, Y.~Wang, J.~Wang, and M.~Long.
\newblock MetaSets: Meta-learning on point sets for generalizable
representations.
\newblock In \emph{CVPR}, 2021.

\bibitem{wang2019dgcnn}
Y.~Wang, Y.~Sun, Z.~Liu, S.~E.~Sarma, M.~M.~Bronstein, and J.~M.~Solomon.
\newblock Dynamic graph CNN for learning on point clouds.
\newblock \emph{ACM Trans.~Graph.}, 2019.

\bibitem{dhariwal2020jukebox}
P.~Dhariwal, H.~Jun, C.~Payne, J.~W.~Kim, A.~Radford, and I.~Sutskever.
\newblock Jukebox: A generative model for music.
\newblock \emph{arXiv:2005.00341}, 2020.

\bibitem{lancucki2020robust}
A.~{\L}a{\'n}cucki, J.~Chorowski, G.~Sanchez, R.~Marxer, N.~Chen,
H.~J.~G.~A.~Dolfing, S.~Khurana, T.~Alum{\"a}e, and A.~Laurent.
\newblock Robust training of vector quantized bottleneck models.
\newblock In \emph{IJCNN}, 2020.
\end{thebibliography}
\end{document}